\documentclass[sigconf]{acmart}

\renewcommand\footnotetextcopyrightpermission[1]{} 
\usepackage{amsfonts}
\usepackage{amssymb,amsbsy}
\usepackage{amsmath}
\usepackage{amsfonts, amscd, amsthm}
\usepackage{algorithm, algorithmic}
\usepackage{color}
\usepackage{url}
\usepackage{verbatim}
\usepackage{setspace}
\usepackage[abs]{overpic}
\usepackage{graphicx}
\usepackage{subfigure}
\usepackage{bbm, bm}
\usepackage{accents}
\usepackage{fullpage}
\usepackage{dsfont}
\usepackage{bbm}

\newtheorem*{example*}{Example}
\numberwithin{equation}{section}

\renewcommand{\bm}{\boldsymbol}
\newcommand{\mc}{\mathcal}

\newcommand{\set}[1]{\left\{ #1 \right\}}

\newlength{\dhatheight}

\usepackage{booktabs} 
\usepackage{multirow}
\usepackage{array}
\usepackage{multicol}
\usepackage{arydshln}

\newcolumntype{L}[1]{>{\raggedright\let\newline\\\arraybackslash\hspace{0pt}}m{#1}}
\newcolumntype{C}[1]{>{\centering\let\newline\\\arraybackslash\hspace{0pt}}m{#1}}
\newcolumntype{R}[1]{>{\raggedleft\let\newline\\\arraybackslash\hspace{0pt}}m{#1}}

\acmDOI{}

\acmISBN{}

\setcopyright{rightsretained}
\acmConference[SIGIR 2019 eCom]{ACM SIGIR Workshop on eCommerce}{July 2019}{Paris, France}
\acmYear{2019}
\copyrightyear{2019}
\makeatletter
\renewcommand\@formatdoi[1]{\ignorespaces}
\makeatother
\acmISBN{}

\usepackage{fmtcount} 
\usepackage{array,multirow}

\newcolumntype{C}[1]{>{\centering\arraybackslash}m{#1}}

\usepackage{hyperref}

\makeatletter
\newcommand{\printfnsymbol}[1]{%
	\textsuperscript{\@fnsymbol{#1}}%
}
\makeatother

\begin{document}
\title[FAISS vs. FENSHSES]{An Empirical Comparison of FAISS and FENSHSES for Nearest Neighbor Search in Hamming Space}

\author{Cun (Matthew) Mu\printfnsymbol{2}}
\thanks{\printfnsymbol{2} C. Mu and B. Yang contributed equally to this work}
\affiliation{%
	\institution{Walmart Labs}
	\city{Hoboken}
	\state{NJ}
}
\email{matthew.mu@jet.com}

\author{Binwei Yang\printfnsymbol{2}}
\affiliation{%
	\institution{Walmart Labs}
	\city{Sunnyvale}
	\state{CA}
}
\email{BYang@walmartlabs.com}

\author{Zheng (John) Yan}
\affiliation{%
	\institution{Walmart Labs}
	\city{Hoboken}
	\state{NJ}
}
\email{john@jet.com}

%
%
%

\renewcommand{\shortauthors}{C. Mu et al.}

\begin{abstract}
In this paper, we compare the performances of FAISS and FENSHSES on nearest neighbor search in Hamming space--a fundamental task with ubiquitous applications in nowadays eCommerce. Comprehensive evaluations are made in terms of indexing speed, search latency and RAM consumption. This comparison is conducted towards a better understanding on trade-offs between nearest neighbor search systems implemented in main memory and the ones implemented in secondary memory,  which is largely unaddressed in literature.

%
%
\end{abstract}

%
%

%




\maketitle

\section{Introduction}
Nearest neighbor search (NNS) within semantic embeddings (a.k.a., vector similarity search) has become a common practice in ubiquitous eCommerce applications including neural ranking model based text search \cite{Brenner2018ete, magnani2019neural}, content-based image retrieval  \cite{yang2017visual, mu2018towards},  collaborative filtering \cite{deshpande2004item}, large-scale product categorization \cite{hu2018best}, fraud detection \cite{raghava2017predicting}, etc. While vector similarity search is capable of substantially boosting search relevancy by understanding customers' intents more semantically, it presents a major challenge: how to conduct nearest neighbor search among millions or even billions of high-dimensional vectors in a real-time and cost-effective manner. 

The fundamental trade-off between search latency and cost-effectiveness would naturally classify nearest neighbor search solutions into two broad categories.

 \paragraph{NNS solutions implemented in main memory.} 
 This type of NNS solutions has been extensively studied and explored in the field of information retrieval (IR).  As a result, the majority of those widely used ones (e.g., Spotify's Annoy \cite{Github:annoy}, Facebook's FAISS \cite{johnson2017billion} and  Microsoft's SPTAG \cite{wang2012scalable, ChenW18}) in nowadays software market fall into this category. 

 \paragraph{NNS solutions implemented in secondary memory.}
In contrast, the second type of NNS solutions are delivered only recently by active efforts from both academia and industry \cite{lux2013visual,rygl2017semantic,ruuvzivckaflexible,amato2018large,mu2018towards, mu2019empowering} to empower full-text search engines (e.g., Elasticsearch and Solr) with the capability of finding nearest neighbors. By leveraging inverted-index-based information retrieval systems and cutting-edge engineering designs from these full-text search engines, such full-text search engine based solutions are capable of economically reducing RAM consumption \cite{amato2018large}, coherently supporting multi-modal search \cite{mu2018towards} and being extremely well-prepared for production deployment \cite{rygl2017semantic}. However, some of the critical performance questions have not been quantitatively answered in literature:
\begin{itemize}
	\item how much RAM could these full-text search based solutions save?
	\item how much search latency would these solutions sacrifice in order to reduce RAM consumption?
\end{itemize}

In this paper, we will shed light on the above questions through a case study on the task of nearest neighbor search in Hamming space (i.e., the space of binary codes). This task is an extremely important subclass of NNS, as learning and representing textual, visual and acoustic data
with compact and semantic binary vectors is a pretty mature
technology and common practice in nowadays IR systems. In particular, eBay recently builds its whole visual search system \cite{yang2017visual}  upon finding nearest neighbors within binary embeddings generated through deep neural network models.

We choose one representative solution of each category--FAISS (Facebook AI Similarity Search) from Facebook's AI Research Lab \cite{johnson2017billion} and FENSHSES (Fast Exact Neighbor Search in Hamming Space on Elasticsearch) from the search and catalog teams at Walmart Labs \cite{mu2019empowering, mu2019empowering2}--to evaluate their performances in finding nearest neighbors within binary codes.

\section{FAISS \lowercase{vs.} FENSHSES}
We will compare performances of FAISS and FENSHSES from three key perspectives: time spent in data indexing, search latency and RAM consumption.

\paragraph{Data generation.} Our dataset $\mc B$ is generated using 2.8 million images selected from Walmart.com's home catalog through pHash \cite{klinger2010phash, christoph2010implementation}--one of the most effective perceptual hash schemes in generating fingerprints for multimedia files (e.g. images, audios and videos)--with number of bits $m\in \set{64, 256, 1024, 4096}$ respectively.  Note that vector similarity search based on pHash has been widely used in a variety of visual tasks including forensic image recognition \cite{peter2012privacy}, duplicate image detection \cite{chaudhuri2018smart} and copyright protection \cite{mehta2019decentralised}, etc.

\paragraph{Settings.} 
For FAISS, we use its binary flat index with five threads. For a fair comparison, we accordingly deploy FENSHSES by creating its Elasticsearch index with five shards and zero replica. The rest of configurations are left as their default and suggested values.  Both FAISS and FENSHSES are set up and tested on the same Microsoft Azure virtual machine.

\paragraph{Speed in indexing.} During the indexing phase, FAISS indexes the data into main memory (i.e., RAM), while FENSHSES indexes the data into secondary memory (e.g., hard disk). As a consequence, FAISS is much faster than FENSHSES in terms of data indexing (see Table \ref{tab: speed_in_indexing}). But on the other hand, whenever the process is killed and needs a restart, FAISS has to go through this procedure again to re-index data into RAM, while FENSHSES could unaffectedly use its built index on hard disk without re-indexing.

\begin{table}[h!]
	\begin{tabular}{|C{1.6cm}|C{1.6cm}|C{1.5cm}|C{1.5cm}|}
		\hline 
		\textbf{\# of Bits} & \textbf{FAISS (sec.)} & \textbf{FENSHSES (sec.)}
		\\
		\hline  
		\hline
		{64} & 18.5 & 75.5 \\
		{256} & 37.7 & 140.2 \\
		{1024} & 111.9 & 369.5 \\
		{4096} & 397.3 & 1300.9 \\
		\hline
	\end{tabular}
	\caption{{\bf Indexing time consumption.} {\normalfont  FAISS is about four times faster than FENSHSES in creating the index for nearest neighbor search. }}\label{tab: speed_in_indexing}
\end{table}

\paragraph{Search latency.} We randomly select $10,000$ binary codes from $\mc B$ to act as query codes. For each query code $\bm q$, we instruct  FAISS and FENSHSES to find all $r$-neighbors of $\bm q$ in $\mc B$, namely \begin{flalign}
B_H(\bm q, r) := \set{\bm b \in \mc B \;\vert\; d_H(\bm b, \bm q) \le r},
\end{flalign}
where $d_H(\bm b, \bm q):=\sum_{i=1}^m  \mathbbm 1_{\set{b_i \neq q_i}}$ denotes the Hamming distance between binary code $\bm b$ and $\bm q$, and the Hamming radius $r\ge 0$. As shown in Table 2, FENSHSES is quite competitive for small radius $r$. This is because FENSHSES fully leverages Elasticsearch's inverted index to first conduct a sub-code filtering to only consider a subset of $\mc B$ for Hamming distance computation, which is most effective for small $r$. In contrast, FAISS scans every binary code in $\mc B$, so its search latency is almost invariant with respect to $r$. For applications (e.g., near-duplicate image detection and visual search) where we care most about nearest neighbors within a small radius, FENSHSES could be in a more favorable position than FAISS.

\paragraph{RAM consumption.} 
Since FAISS is implemented in main memory, its RAM consumption undoubtedly rises along with the increase in the size of dataset $\mc B$, as shown in Table \ref{tab: RAM}.  In contrast, by leveraging the highly optimized disk-based index mechanics behind full-text search engines, FENSHSES consumes a much smaller amount of RAM when conducting nearest neighbor search. This property makes FENSHSES more cost-effective and thus  more suitable especially to big-data applications. 

\begin{table}[h!]
	\begin{tabular}{|C{1.6cm}|C{1.6cm}|C{1.5cm}|C{1.5cm}|}
		\hline 
		\textbf{\# of Bits} & \textbf{$r$} & \textbf{FAISS (ms)} & \textbf{FENSHSES (ms)}
		\\
		\hline  
		\hline
		\multirow{3}{*}{64} & 3 & 34.0 & \textbf{5.8} \\ 
		& 7& 37.0 & \textbf{25.7} \\ 
		& 11& \textbf{42.7} & 117.7 \\ 
		\hline
		\multirow{3}{*}{256} & 15 & 42.9 & \textbf{7.8} \\ 
		& 31& 42.8 & \textbf{22.5} \\ 
		& 47& \textbf{45.4} & 77.7 \\ 
		\hline
		\multirow{3}{*}{1024} & 63 &79.2 & \textbf{31.6} \\ 
		& 127& \textbf{81.9} & 89.9 \\ 
		& 191& \textbf{90.4} & 250.0 \\ 
		\hline
		\multirow{3}{*}{4096} & 255 & 222.7& \textbf{134.2} \\ 
		& 511& \textbf{223.2} & 612.5 \\ 
		& 767& \textbf{223.3} & 1797.5 \\ 
		\hline
	\end{tabular}
	\caption{{\bf Search latency.}  {\normalfont  FENSHSES is quite competitive for $r$-neighbor search when the Hamming distance $r$ is small, while the performance of FAISS is pretty robust with respect to $r$. This provides FAISS and FENSHSES different edges for the task of NNS.}  \label{tab: search_latency}}
\end{table}

\begin{table}[h!]
	\begin{tabular}{|C{1.6cm}|C{1.6cm}|C{1.5cm}|C{1.5cm}|}
		\hline 
		\textbf{\# of Bits} & \textbf{$r$} & \textbf{FAISS (GB)} & \textbf{FENSHSES (GB)}
		\\
		\hline  
		\hline
		\multirow{3}{*}{64} & 3 & 2.2 & 1.6 \\ 
		& 7& 2.2 & 1.6\\ 
		& 11& 2.2 & 1.6 \\ 
		\hline
		\multirow{3}{*}{256} & 15 & 2.3 & 1.6 \\ 
		& 31& 2.3 & 1.6 \\ 
		& 47& 2.3 & 1.6\\ 
		\hline
		\multirow{3}{*}{1024} & 63 & 2.9 & 1.6 \\ 
		& 127& 2.9 & 1.6 \\ 
		& 191& 2.9 & 1.6 \\ 
		\hline
		\multirow{3}{*}{4096} & 255 & 4.9 & 1.6 \\ 
		& 511& 4.9 & 1.6 \\ 
		& 767& 4.9 & 1.6 \\ 
		\hline
	\end{tabular}
	\caption{{\bf Main memory (RAM) consumption.} {\normalfont  The RAM consumed by FAISS substantially grows with the increase in the size of dataset $\mc B$. In contrast, FENSHSES consumes a constant amount of  RAM, which is much smaller than the one consumed by  FAISS.}}  \label{tab: RAM}
\end{table}

\section{Conclusion}
In this case study, we compare FAISS and FENSHSES for the task of nearest neighbor search in Hamming space. By evaluating their performances in terms of speed in data indexing, search latency and RAM consumption, we hope practitioners could now better understand the pros and cons of the main memory based NNS solutions and the secondary memory based ones, and thus make their best choices accordingly (at least in NNS systems within binary cods). In the future, we will compare FAISS and FENSHSES under a wider range of applications;  and moreover, we will also go beyond Hamming space to evaluate vector similarity search systems for general NNS problems. 

\section*{Acknowledgement}
We are grateful to three anonymous reviewers for their helpful
suggestions and comments that substantially improve the paper.   CM would like to thank Jun Zhao and Guang Yang for insightful discussions on FENSHSES. BY would like to thank Alessandro Magnani for helpful discussions on pHash and its related applications, and Zuzar Nafar for his support on this study. 
 


\bibliographystyle{ACM-Reference-Format}
\bibliography{vss}
\end{document}